\documentclass[a4paper]{article}

\usepackage{INTERSPEECH2020}

\usepackage[normalem]{ulem}
\usepackage{amsmath}
\usepackage{multirow}

\usepackage{latexsym}
\usepackage[hang,flushmargin]{footmisc}

\usepackage{amsfonts}
\usepackage{algorithm}
\usepackage{algorithmic}
\usepackage{graphicx}
\usepackage{subcaption}
\usepackage{booktabs}

\title{Improving End-to-End Speech-to-Intent Classification with Reptile}
\name{Yusheng Tian\textsuperscript{*}, Philip John Gorinski
\thanks{$^*$Work conducted while this author was an intern at Huawei Noah's Ark Lab, London.}}

\address{Huawei Noah's Ark Lab, London, UK }
\email{yusheng.tian@hotmail.com, philip.john.gorinski@huawei.com}

\begin{document}
	
	\maketitle
	\begin{abstract}
		End-to-end spoken language understanding (SLU) systems have many advantages over conventional pipeline systems, but collecting in-domain speech data to train an end-to-end system is costly and time consuming. One question arises from this: how to train an end-to-end SLU with limited amounts of data? Many researchers have explored approaches that make use of other related data resources, typically by pre-training parts of the model on high-resource speech recognition. In this paper, we suggest improving the generalization performance of SLU models with a non-standard learning algorithm, Reptile. Though Reptile was originally proposed for model-agnostic meta learning, we argue that it can also be used to directly learn a target task and result in better generalization than conventional gradient descent. In this work, we employ Reptile to the task of end-to-end spoken intent classification. Experiments on four datasets of different languages and domains show improvement of intent prediction accuracy, both when Reptile is used alone and used in addition to pre-training. 
		
	\end{abstract}
	
	\section{Introduction}
	
	Spoken language understanding (SLU) is a key component in assistive conversational agents. The goal of SLU is to infer users’ intentions from speech utterances, such that actions can be taken accordingly to meet users’ requests \cite{wang2005spoken, tur2010left}. Conventional SLU is a pipeline of automatic speech recognition (ASR) and natural language understanding (NLU). End-to-end SLU \cite{1stend2end, serdyuk2018towards, haghani2018audio}, on the other hand, directly maps audio to semantics without the intermediate ASR. Compared to the conventional pipeline SLU, it avoids error propagation, requires less computation, and has the potential to utilize information that is only present in speech but not in text.
	
	Despite these advantages, training end-to-end SLU systems usually requires in-domain annotated audio data, which is very expensive and time-consuming to collect. Due to time and cost constraints, even for high-resource languages like English, current SLU datasets usually only contain less than 20 hours of speech data. Models trained on such limited data are at risk of over-fitting and may generalize poorly on unseen cases, e.g. a new speaker or a paraphrased command. This has motivated researchers to explore approaches that leverage other related data resources, typically by pre-training parts of the model using a high-resource ASR \cite{chen2018spoken, CTCend2end2, Lugosch2019}, whose language might even be different from that of the SLU, when the SLU target language itself is of low-resource \cite{Bhosale2019}. Other approaches include multi-task learning with ASR if audio transcription data is available \cite{haghani2018audio}, and training SLU on synthetic speech \cite{lugosch2019using}. While these methods all prove to be effective, we propose to improve SLU \mbox{models'} generalization capabilities by training with a non-standard learning algorithm: Reptile \cite{Reptile}. This allows us to improve the model performance when no additional data is available, or achieve a further improvement on top of pre-training.
	
	Reptile was originally proposed as a first-order algorithm for model-agnostic meta-learning (MAML) \cite{finn2018learning}. Like the classic MAML, Reptile is well-established in few-shot learning and can optimize generalization, which is very desirable in low-resource settings. However, it can't be directly applied to end-to-end SLU: the original intention of Reptile is to learn a good parameter initialization from multiple source tasks (e.g. image classification of different categories) for fast adaption on a new but related task (e.g. image classification of a new group of categories), but for SLU we lack such source tasks. However, the formulation of Reptile gives rise to an interesting research question: Can we make use of Reptile to optimize generalization during the model training phase, instead of using it as a model initializer? In this paper, we argue that Reptile can be adapted to directly learn an SLU task by dropping the “task sampling” procedure in its original algorithm, resulting in better generalization than conventional gradient descent. We describe analysis justifying this argument by comparing the meta-gradients of the classic MAML and Reptile. We apply Reptile to end-to-end speech-to-intent classification and test its efficacy on 4 datasets of different languages and domains. Experiment results on all datasets show improvement of intent classification accuracy, both when Reptile is used alone and with pre-training.
	
	The contributions of this paper are as follows: (i) we motivate the use of Reptile for single-task learning in low-resource settings; (ii) we adapt and employ Reptile to the task of end-to-end speech-to-intent classification; (iii) we show how our method helps boost performance on diverse datasets of 4 different languages and domains, both with and without pre-training.

	\section{Modeling End-to-End SLU}
	In this work we concentrate on end-to-end speech-to-intent classification. Given a speech command, we would like to find the most probable intent label from pre-defined categories. For example, in smart-home setting, the utterance \textit{``Turn the heat down''} would be mapped to the intent \texttt{decrease\_heat}. Almost invariably, current end-to-end models achieve this goal in two steps. First, an encoder maps the input sequence of audio signals (pre-processed acoustic features or raw waveforms) to a fixed-length utterance embedding. Then a decoder predicts a probability distribution over all possible intent labels conditioned on this utterance embedding. The intent label with the highest probability is selected as the output. 
	
	There has been some work on improving intent classification by utilizing a novel architecture: \cite{CapsuleSLU} replaced the soft-max classifier with a capsule network, and showed that it can make efficient use of limited training data. However, their model is a speaker-dependent system and makes use of pre-defined speech commands; \cite{architecture} used a multi-label classifier instead of a single-label classifier for intent prediction, but this architecture is tailored to a certain type of datasets whose intents are combinations of several slots.
	
\begin{figure}
		\centering
		\includegraphics[width=0.6\linewidth]{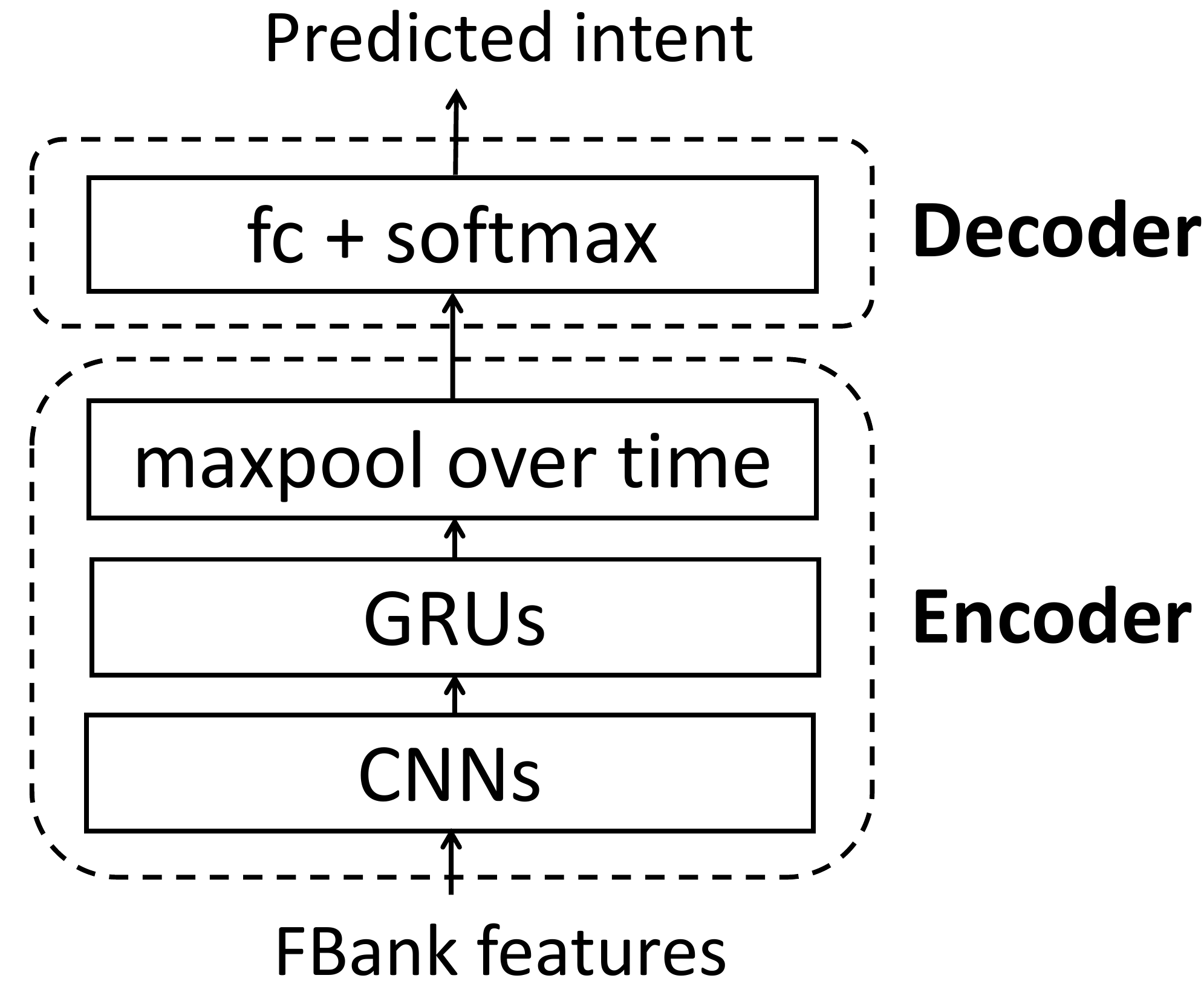}
		\caption{Encoder-decoder architecture for SLU}
		\label{fig:architecture}
	\end{figure}

	Since our main focus in this paper is on the learning algorithm rather than the model architecture, we adopt a simple encoder-decoder architecture similar to that in \cite{serdyuk2018towards} and \cite{Bhosale2019}, illustrated in Figure \ref{fig:architecture}. The choice of a simple architecture also ensures that when comparing our models with SotA results -- see section 5 -- the relative gain of intent prediction accuracy comes from the training strategy rather than a more advanced architecture. Our model architecture is not restricted to a certain type of datasets, and can be augmented for experiments with pre-training, for example by replacing the bottom CNN layers with pre-trained ASR layers, as in \cite{Bhosale2019}.  Details of the model architecture and implementation are described in Section 4.
	
	\section{Reptile Learning}
	Reptile \cite{Reptile} is proposed as a first-order algorithm for MAML \cite{finn2018learning, FinnICML}, which aims to train a model on multiple source tasks, such that the resulting model can be fine-tuned on a new but related task with only a small amount of training examples. In other words, MAML learns an initialization, rather than a good model \cite{gu-etal-2018-meta}. We argue that while this applies to the classic MAML algorithm, Reptile can be adapted to learn a low-resource target task (in our case, spoken intent classification) and improve the model's generalization performance. To explain why this works, we will first give a brief description of Reptile and its origin, the classic MAML, and then introduce the adapted Reptile for SLU.
	
	\subsection{Reptile as a MAML algorithm}
	The classic MAML accomplishes the meta-learning goal by directly incorporating the fast adaptation process into its objective function:
	
	\begin{equation}
	\min_\theta\mathbb{E}_{\tau\sim p(\mathrm{T})} \left[L_\tau\left (U_\tau ^k \left(\theta, D_\tau^{tr} \right ), D_\tau^{test} \right ) \right ]
	\label{eqn:maml}
	\end{equation}
	
	We can interpret the objective as: for any task $\tau$ that is sampled from a distribution of tasks $p(\mathrm{T})$, search for an initialization of model parameters $\theta$, such that after $k$ gradient descent updates (e.g. SGD, Adam) on the training set $D_\tau^{tr}$ (usually of a small size), the model with updated parameter vector $U_\tau ^k \left(\theta, D_\tau^{tr} \right )$ has a minimum loss $L_\tau$ on the test set $D_\tau^{test}$. In other words, we are using $U_{\tau}^{k}$ to find a good \emph{initialization} of $\theta$ for future training, but we are not training $\theta$ itself.
	Also note that $D_\tau^{tr}$ and $D_\tau^{test}$ in the objective function (\ref{eqn:maml}) are different from the conventional training and test set. They refer to the meta training and test sets, which form the whole training data.
	
	MAML solves the above optimization problem through stochastic gradient descent, i.e. by repeatedly sampling a task $\tau$ and updating $\theta$ with 
	\begin{equation}
	\theta \leftarrow \theta - \alpha g_{MAML}
	\label{eqn:mamlupdate}
	\end{equation}
	where $\alpha$ is the meta step size, and
	\begin{equation}
	g_{MAML}=\nabla_\theta\left[L_\tau\left (U_\tau ^k \left(\theta, D_\tau^{tr} \right ), D_\tau^{test} \right ) \right ]
	\end{equation}
	
	The above meta update involves a second-order gradient and might be computationally expensive. Reptile simplifies the meta gradient as (\ref{eqn:reptile}), which is very convenient to compute:
	
	\begin{equation}
	g_{Reptile}=\theta-U_\tau^{k}\left (\theta, D_\tau \right )
	\label{eqn:reptile}
	\end{equation}
	and similar to (\ref{eqn:mamlupdate}), we now update model parameters $\theta$ with
	\begin{equation}
	\theta \leftarrow \theta - \alpha g_{Reptile} = \theta + \alpha (U_\tau^{k}\left (\theta, D_\tau \right )-\theta)
	\label{eqn:reptileupdate}
	\end{equation}

	\subsection{Reptile for SLU}
\begin{figure}
		\centering
		\begin{subfigure}{.2\textwidth}
			\centering
			\includegraphics[width=\linewidth]{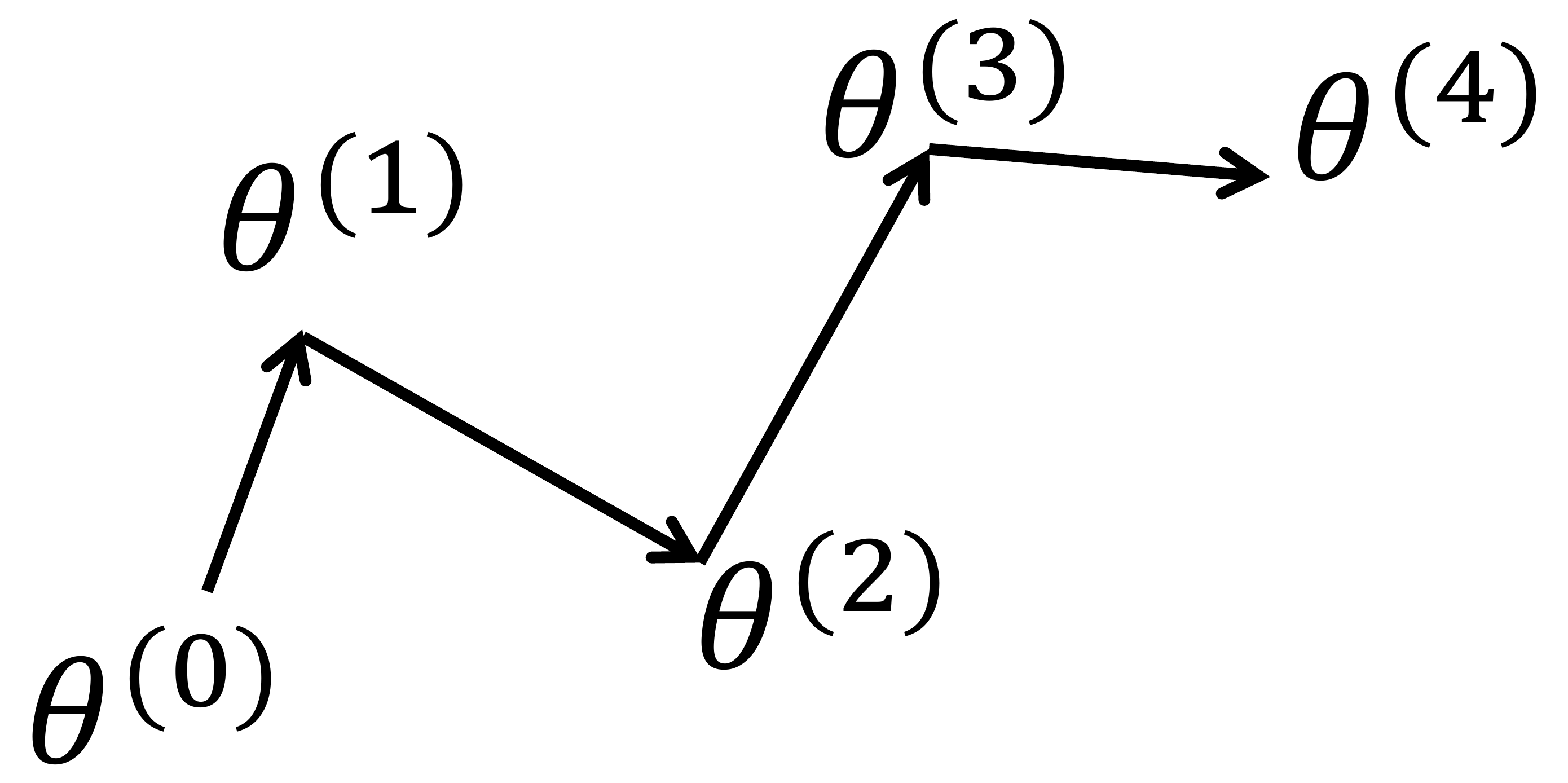}
			\caption{without Reptile}
			\label{fig:sub1}
		\end{subfigure}%
		\begin{subfigure}{.2\textwidth}
			\centering
			\includegraphics[width=\linewidth]{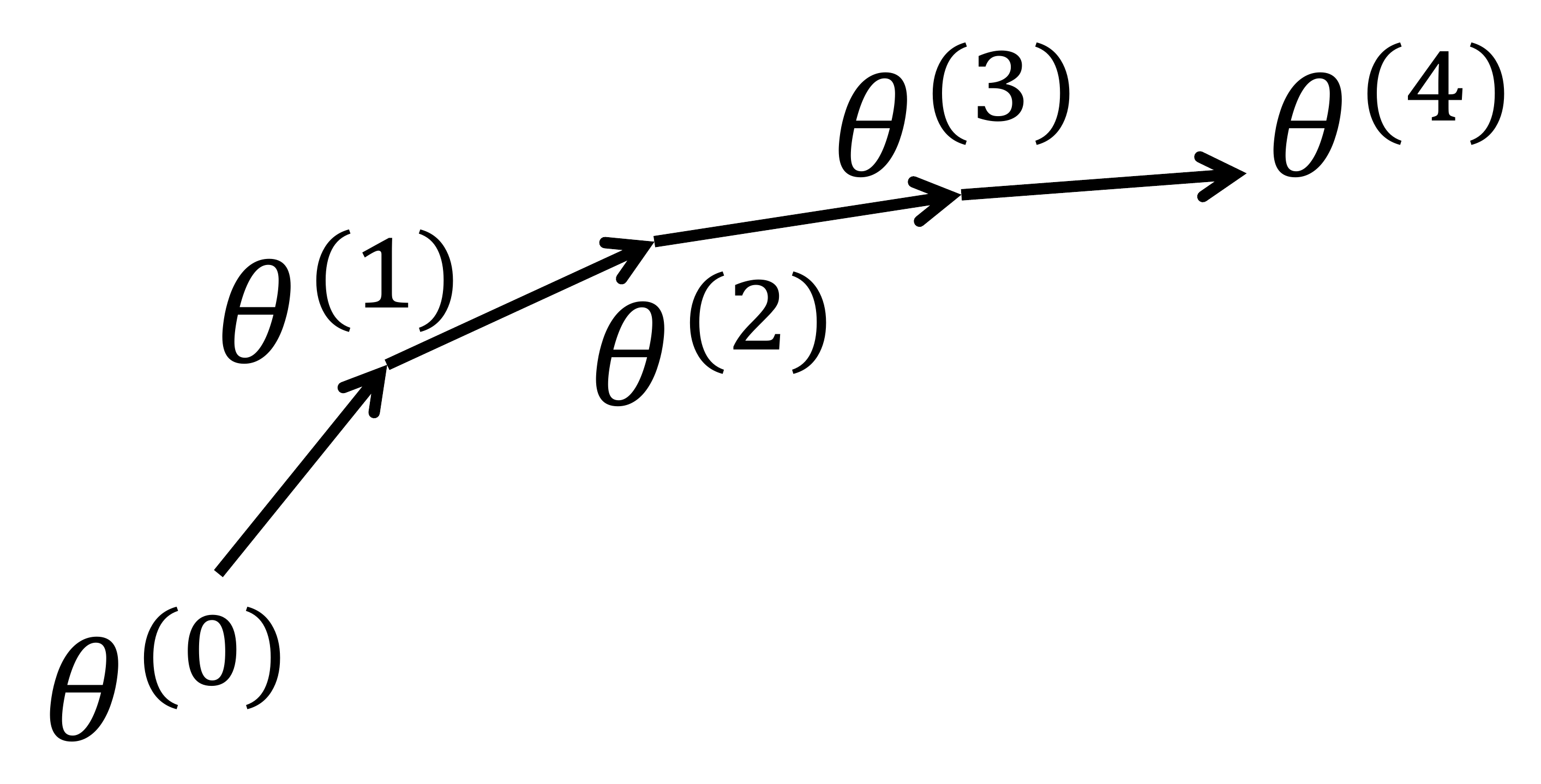}
			\caption{with Reptile}
			\label{fig:sub2}
		\end{subfigure}
		\caption{An illustration of Reptile encouraging gradients computed at different steps to point to similar directions.}
		\label{fig:Ksteps}
	\end{figure}

	We can see from (\ref{eqn:reptile}) that Reptile entirely eliminates the train/test split for meta optimization, therefore there is not a meaningful objective function as in (\ref{eqn:maml}) that corresponds to the meta gradient of Reptile. This means that, as opposed to classic MAML, Reptile is not strictly defined for learning a model initialization. In addition, from (\ref{eqn:reptileupdate}) we can see that Reptile still pushes model parameters towards the trained weights of standard (e.g. SGD, Adam) training algorithms, just at a slower pace. This suggests that unlike the classic MAML, the resulting model of Reptile learning is still a good model itself for the tasks that it has been trained on. On the other hand, like the classic MAML, the meta gradient of Reptile ($g_{Reptile}$) contains some terms that maximize the inner product between gradients computed at different steps (e.g. different mini-batches). This in turn encourages gradients at different steps to point to similar directions, as illustrated in Figure \ref{fig:Ksteps}. We refer readers to \cite{Reptile} for a detailed theoretical proof of this. By encouraging gradients at different steps to point to similar directions, Reptile promotes within-task generalization. This is very beneficial for training SLU with only limited amounts of data. 
	
	When we apply Reptile to training end-to-end SLU, we only learn a single task. Therefore there is no need to repeatedly sample a task like the original Reptile, and the algorithm can be simplified as Algorithm 1. We have eliminated the subscript $\tau$ because it always refers to the same task of spoken intent prediction. The basic unit of iteration in Reptile training are Episodes (operations within the While loop, Algorithm1), comprising k Epochs training plus one interpolation. Details of the training procedure are described in Section 4.

	Our adapted Reptile algorithm is very similar to that of the Lookahead Optimizer in \cite{lookahead}. The Lookahead algorithm is not based on Reptile, but also contains $k$ steps forward and $1$ interpolation step. The major difference between these two methods is that in the adapted Reptile, the interpolation happens over the full data ($k$ epochs), while the Lookahead Optimizer operates on a much smaller scale (around 10 mini-batches). The Lookahead optimizer also does not target low-resource tasks, which is a direct motivation for our adaptation of Reptile.

	\begin{algorithm}[t]
		\caption{Reptile for end-to-end SLU}
		\begin{algorithmic}[1]
			\REQUIRE{$k$: number of inner epochs}
			\REQUIRE{$\alpha$: step size}
			\STATE Randomly initialize $\theta$
			\WHILE{not done}
			\STATE{Compute $\theta^{'}=U ^k \left(\theta, D_{train} \right )$, denoting $k$ gradient descent (SGD/Adam) updates over the entire training set $D_{train}$}, i.e. $k$ epochs.
			\STATE{Update $\theta \leftarrow \theta + \alpha (\theta^{'}-\theta )$}
			\ENDWHILE
		\end{algorithmic}
	\end{algorithm}
	
	\section{Experimental Setup}
	\subsection{Datasets}
	We use the following four SLU datasets. (1) fluent speech commands (FSC) \cite{Lugosch2019}: English speech commands related to personal assistant services. (2) Grabo \cite{Grabo}: speech commands in Dutch for controlling robot movement. (3) Tamil \cite{karunanayake-etal-2019-transfer}: speech commands in the low-resource language Tamil for requesting banking services. (4) CN: internally collected Mandarin Chinese speech commands for operating a mobile phone. Audio from different speakers was recorded using the same mobile phone. Statistics of the four datasets are summarized in Table \ref{tab:dataset}.
	
	\begin{table}
		\caption{Statistics of the SLU datasets. $^*$ indicates the data is for each speaker.}
		\label{tab:dataset}
		\centering
		\begin{tabular}{l|rrrr}
			\toprule
			Dataset & FSC & Grabo & Tamil & CN\\
			\midrule
			\#intents & 31 & 36$^*$ & 6 & 130\\
			\#phrases & 248 & 36$^*$ & 31 & 13756 \\
			\#utts & 30043 & 540$^*$ & 400 & 14771\\
			\#speakers & 97 & 10\hspace{1ex} & 40 & 42 \\
			\bottomrule
		\end{tabular}
	\end{table}
	
	We use the default train/validation/test split of the FSC dataset. The Grabo dataset is designed for developing speaker-dependent systems, with each speaker having recorded 36 different commands with 15 repetitions. For each speaker, we randomly select 2 recordings of each command for training, 4 for validation, and use the remaining 9 recordings for testing. On the Tamil dataset we perform 5-fold cross-validation  as in \cite{karunanayake-etal-2019-transfer}. We split the dataset into 5 parts, such that each subset has an approximately equal number of samples, and no speaker is spread across different subsets. Each fold is repeated 3 times to allow different train/validation combinations. For the CN dataset we have 6 main speakers. Each of them has contributed over 1200 recordings. We use recordings from these speakers as the training data. We randomly select 24 out of the remaining 36 speakers and use their recordings (around 4500 audio clips) as the held-out set for testing. The rest of the data (around 2400 audio clips) forms the validation set.
	
	\subsection{Model architecture and training}
	Our model takes 39-dimensional log Mel-Filterbank feature plus energy as input, computed with a window size of 25 ms and a shift of 10 ms. The encoder consists of a stack of 2-layer \mbox{2-D} CNN  and 2-layer uni-directional GRU. For each CNN layer, we stride with a factor of 2 along time to reduce the sequence length, and apply batch normalization as well as dropout. For all datasets we use GRU cells of 128 hidden units, except for the CN dataset, where we set the number to 256 as it has significantly more target intents. The output of the GRU layers is fed into a maxpooling layer over all time steps in order to extract fixed-length utterance embeddings. The decoder is a simple softmax classifier with only 1 hidden layer.
	
	We investigate the effectiveness of Reptile both when the model is trained from scratch, and when utilizing pre-training. For pre-training, we follow the strategy proposed in \cite{Bhosale2019} which incorporates layers extracted from pre-trained acoustic models to the input of SLU models, and supports cross-lingual transfer learning. For all SLU datasets, regardless of the spoken language, we use the hidden representation of a pre-trained end-to-end English phoneme recognizer provided in the source code of \cite{Lugosch2019} as our model's input, and directly feed it to the GRU layers. There is no need to keep the CNN layers in our model when pre-training is applied, because all its functionality can be achieved by the pre-trained ASR layers.
	
	We use the Adam optimizer \cite{Adam} with a learning rate of 0.001 both when training with Reptile and without. For Reptile learning, we set the inner epoch iteration to $k=5$ for all datasets. We empirically set a step size $\alpha=0.3$ for Grabo, and $\alpha=0.1$ for all other datasets. Within each epoch, we train our model on mini-batches of data, and experiment results show that using a small batch size helps stabilize training when we have extremely low-resource data (e.g. we used a batch-size of 8 for the Grabo and Tamil dataset). We train with Reptile on the training set. After each episode, we record the intent prediction accuracy on the validation set to examine model's generalization performance, and decide whether to stop training. During each training, we employ early stopping with patience of 10. 
	
	\section{Results}
	\subsection{Effectiveness of Reptile}
	Table \ref{tab:results} shows the intent classification accuracies of our models under all training conditions for all four SLU datasets, as well as state-of-the-art results for the three freely available datasets. Our results are the average of 10  speakers for the Grabo dataset, the average of 5 folds for the Tamil dataset, and the average of 3 runs for the FSC and CN datasets. It can be seen that Reptile gives improvement on all datasets, both with and without pre-training. The lowest accuracy, but also the most significant improvement is observed on the Chinese dataset, perhaps because this dataset contains over 10,000 different expressions in total while only having 6 speakers in the training set. Note that it is not strictly fair to compare our results with those in the literature because they used different and often more elaborate architectures (e.g. \cite{karunanayakesinhala} used a simple 2-layer CNN encoder, while \cite{CapsuleSLU} used a capsule network for classification), but we can see that with Reptile learning our model with its simple architecture achieves at least comparable intent classification accuracy to state-of-the-art results on the open datasets. The SotA result for the Tamil dataset without pre-training is confusingly low. This is because the number reported in the original paper is from an SVM model, and the authors did not employ a deep-learning model without pre-training.
	
	To assess the significance of improvements gained through Reptile training on the small tested datasets, we carried out two-tailed paired t-tests on the multi-run experiment results of our simple model when trained with and without Reptile. The p-value results on all datasets are within 0.05, as summarized in Table \ref{tab:ttest}, showing that the performance difference between the models trained with Reptile and the baseline is statistically significant. Since we do not have predictions of the SotA systems in all settings, and our focus is on the efficacy of Reptile when compared to standard training, we do not make claims about statistical differences between our best models and SotA results.
	
	\begin{table}
		\caption{Intent prediction accuracy. \textbf{--pretrain} indicates no pre-training is applied, \textbf{+pretrain} indicates using pre-trained ASR layers. Rows labeled \textbf{Base} are for models trained with standard Adam optimization, while \textbf{Reptile} employs our adapted Reptile training. Where applicable, \textbf{SotA} shows previously published State of the Art results as reported in {\footnotesize \textsuperscript{1}\cite{architecture}, \textsuperscript{2}\cite{CapsuleSLU}, \textsuperscript{3}\cite{karunanayake-etal-2019-transfer}, \textsuperscript{4}\cite{lugosch2019using}, \textsuperscript{5}\cite{karunanayakesinhala}}, and \textsuperscript{*} indicates that results were read off a graph.}
		\label{tab:results}
		\centering
		\setlength{\tabcolsep}{5pt}
		\begin{tabular}{ll|cccc}
			\toprule
			& & FSC & Grabo & Tamil & CN\\
			\midrule
			\parbox[t]{1ex}{\multirow{3}{*}{\rotatebox[origin=c]{90}{--pretrain}}} & Base & 98.3 & 88.9 & 85.8 & 25.8\\
			& Reptile & 98.8 & 94.4 & 90.2 & 36.9\\
			& SotA & \hspace{1ex}98.6\textsuperscript{1} & \hspace{2ex}94.5\textsuperscript{*2} & \hspace{1ex}29.2\textsuperscript{3} & --\\
			\midrule
			\parbox[t]{1ex}{\multirow{3}{*}{\rotatebox[origin=c]{90}{+pretrain}}} & Base & 98.7 & 98.6 & 94.2 & 49.9\\
			
			& Reptile & 99.2 & 98.9 & 94.5 & 55.2\\
			& SotA & \hspace{1ex}99.1\textsuperscript{4} & -- & \hspace{1ex}81.7\textsuperscript{5} & --\\
			\bottomrule
		\end{tabular}
		
	\end{table}

\begin{table}
	\caption{Paired t-test results between baseline and Reptile models. \textbf{--pretrain} indicates no pre-training is applied, \textbf{+pretrain} indicates using pre-trained ASR layers. }
	\label{tab:ttest}
	\centering
	\setlength{\tabcolsep}{5pt}
	\begin{tabular}{l|cccc}
		\toprule
		& FSC & Grabo & Tamil & CN\\
		\midrule
		-pretrain & 0.012 & 0.023 & 0.0073 & 0.0051\\
		\midrule
		+pretrain & 0.0042 & 0.026 & 0.014 & 0.0013\\
		\bottomrule
	\end{tabular}
	
\end{table}

	\subsection{Impact of Training Data Size}
\begin{figure}
		\centering
		\includegraphics[width=.95\linewidth]{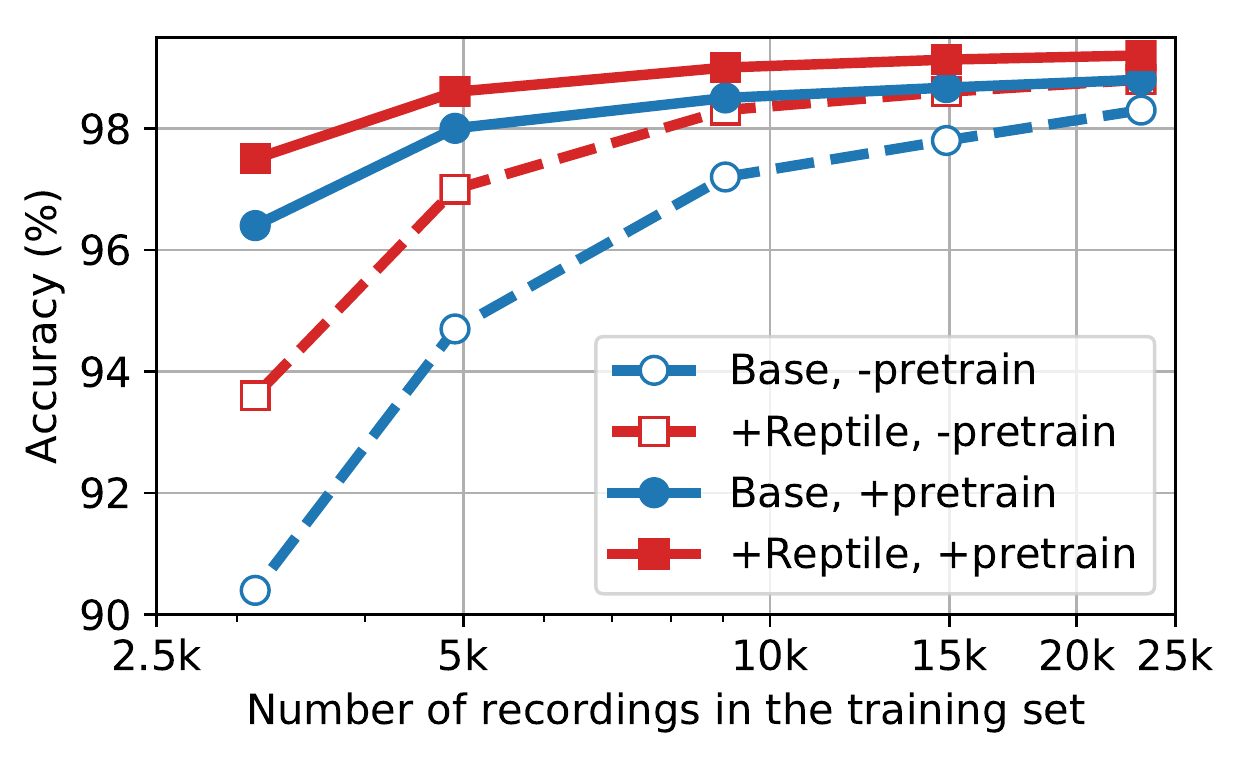}
		\caption{Test set intent prediction accuracy w.r.t the size of training data (FSC dataset, with pre-training).}
		\label{fig:trainsize}
	\end{figure}

	Figure \ref{fig:trainsize} shows the impact of decreasing the amount of training data on intent classification performance for a model trained with and without Reptile. The experiment is carried out on the FSC dataset because it is the only one of the open datasets that is of large enough size. We vary the amount of training data by gradually removing speakers along with their recordings from the training set. We can see that the model trained with Reptile consistently outperforms the baseline model, and the gap between the two models increases as the training data size shrinks. When only $10\%$ of the original data are used for training, the absolute accuracy difference between Reptile and the baseline reaches $1\%$ with pre-training, and over $3\%$ without pre-training. This demonstrates that Reptile learning helps improve the model generalization when there is only limited speaker variation present in the training data. It also shows that under extreme low-resource settings, Reptile can bring appreciable intent prediction improvement even when no pre-training is applied.
	
	\subsection{Learning Curves}
\begin{figure}
		\centering
		\includegraphics[width=.95\linewidth]{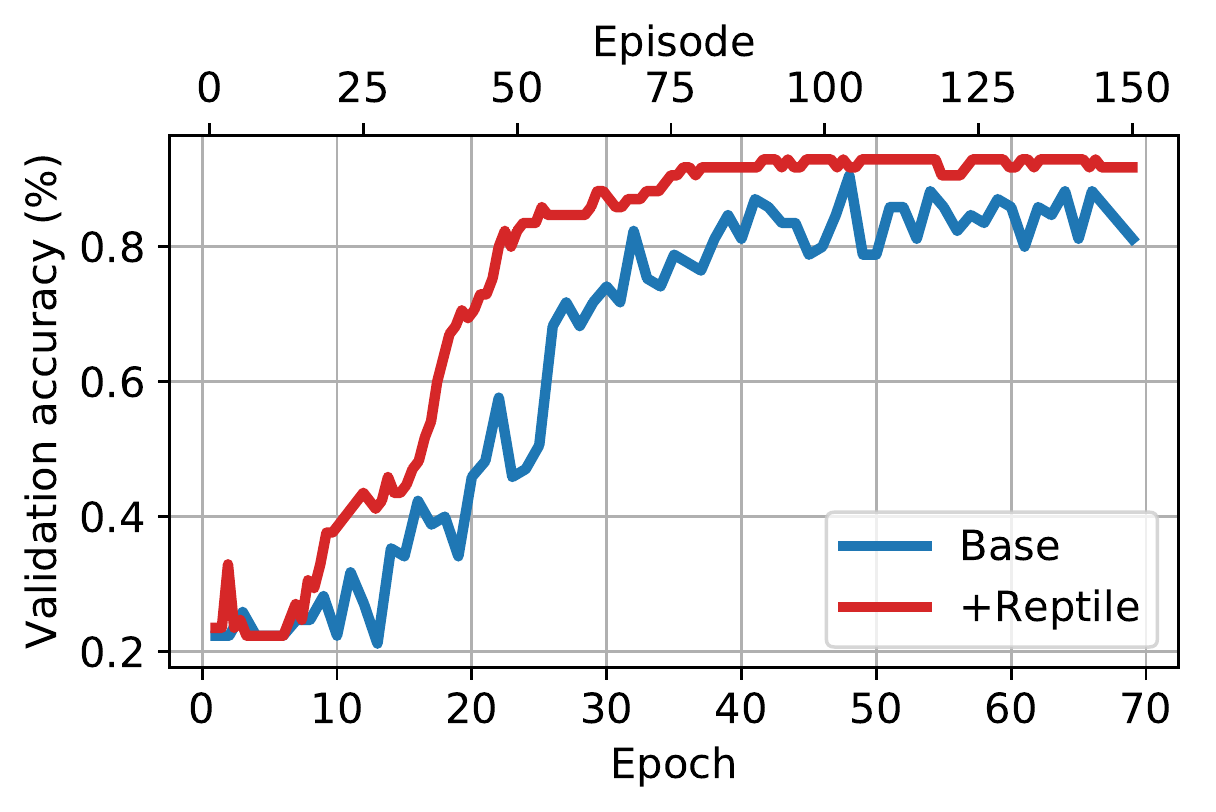}
		\caption{Learning curves of intent prediction accuracy on the validation set (Tamil dataset, w/o pre-training).}
		\label{fig:traincurve}
	\end{figure}

	The difference between Reptile learning and the conventional gradient descent (in our case Adam) is also reflected in their learning curves. Figure \ref{fig:traincurve} compares learning curves of the two optimization approaches on the Tamil dataset, without pre-training. Since the basic unit of iteration in Reptile training is an episode, which consists of $k$ epochs and $1$ interpolation, we cannot compare the $i^{th}$ Reptile episode to the $i^{th}$ epoch of SGD training. But when looking at their learning curves as a whole, we can see that the learning curve of Reptile is much smoother than the learning curve of Adam, with the accuracy on the validation set growing steadily. The model trained with Reptile also reaches a higher steady point of accuracy than the baseline. Similar patterns can also be observed for the other two low-resource datasets: Grabo and CN, sometimes even when pre-training is applied. This conforms with the analysis in Section 3 that Reptile encourages gradients computed at different steps to point to similar directions, promoting generalization.
	
	\section{Conclusions and Future Work}
	In this paper, we applied Reptile to training low-resources, end-to-end SLU, in order to cope with the data scarcity problem. We tested Reptile on various SLU datasets of different languages and domains. Experimental results demonstrate that Reptile improves model generalization, and helps the model to better deal with speaker variations than conventional gradient descent. We should point out that under very low-resource settings pre-training is still the most effective way of bootstrapping model performance, and Reptile can only act as an assistant rather than a replacement of pre-training. In this paper we focus on intent classification. In future work we would like to extend Reptile learning to other SLU tasks such as slot filling.

	\bibliographystyle{IEEEtran}
	
	\bibliography{mybib}
	
\end{document}